\documentclass{article}
\pdfoutput=1

\usepackage{PRIMEarxiv}

\usepackage[utf8]{inputenc} 
\usepackage[T1]{fontenc}    
\usepackage{hyperref}       
\usepackage{url}            
\usepackage{booktabs}       
\usepackage{amsfonts}       
\usepackage{nicefrac}       
\usepackage{microtype}      
\usepackage{lipsum}
\usepackage{fancyhdr}       
\usepackage{graphicx}       
\graphicspath{{media/}}     
\usepackage{amsmath}
\usepackage[hang,flushmargin]{footmisc}
\usepackage{multirow}
\usepackage{graphicx}
\usepackage[font=small,labelfont=bf]{caption}

\usepackage{graphicx}
\usepackage[style=numeric]{biblatex} 
\title{Lightweight Safety Classification Using Pruned Language Models
}

\author{
  Mason Sawtell \\
  Neudesic, an IBM Company \\
  \texttt{mason.sawtell@neudesic.com} \\
  \AND
  Tula Masterman \\
  Neudesic, an IBM Company\\
  \texttt{tula.masterman@neudesic.com} \\
  \And
  Sandi Besen\\
  IBM \\
  \texttt{sandi.besen@ibm.com} \\
  \AND
  Jim Brown \\
  Neudesic, an IBM Company \\
  \texttt{jim.brown@neudesic.com} \\
  \AND
}

\begin{document}
\maketitle

\begin{abstract}

In this paper, we introduce a novel technique for content safety and prompt injection classification for Large Language Models. Our technique, Layer Enhanced Classification (LEC), trains a Penalized Logistic Regression (PLR) classifier on the hidden state of an LLM's optimal intermediate transformer layer. By combining the computational efficiency of a streamlined PLR classifier with the sophisticated language understanding of an LLM, our approach delivers superior performance surpassing GPT-4o and special-purpose models fine-tuned for each task. We find that small general-purpose models (Qwen 2.5 Instruct sizes 0.5B, 1.5B, and 3B) and other transformer-based architectures like DeBERTa v3 are robust feature extractors allowing simple classifiers to be effectively trained on fewer than 100 high-quality examples. Importantly, the intermediate transformer layers of these models typically outperform the final layer across both classification tasks. Our results indicate that a single general-purpose LLM can be used to classify content safety, detect prompt injections, and simultaneously generate output tokens. Alternatively, these relatively small LLMs can be pruned to the optimal intermediate layer and used exclusively as robust feature extractors. Since our results are consistent on different transformer architectures, we infer that robust feature extraction is an inherent capability of most, if not all, LLMs.
\end{abstract}


\let\thefootnote\relax
\footnote{The opinions expressed in this paper are solely those of the authors and do not necessarily reflect the views or policies of their respective employers.}

\section{Introduction}

Since the introduction of LLMs, a primary concern has been detecting inappropriate content in both the user's input and the generated output of the LLM. Establishing language model guardrails is a critical requirement of responsible AI practices. A robust set of guardrails extends beyond the detection of hate speech and sexual content, also detecting when an LLM strays away from its intended purpose. There have been numerous troubling instances where chat bots have been coerced into responding to inappropriate requests and therefore produced damaging outputs. Existing solutions for identifying inappropriate content range in complexity from traditional text classification methods to using an LLM to classify the text. Our approach combines the computational efficiency of a simple machine learning based classifier with the robust language understanding provided by an LLM for optimal performance.

Two primary concerns related to LLM use include content safety and prompt injection detection. Content safety involves identifying inputs and outputs that are harmful, offensive, or otherwise inappropriate \cite{ghosh_aegis_2024, inan_llama_2023, zheng_lmsys-chat-1m_2024, li_salad-bench_2024}. Prompt injections are attempts by the user to manipulate the language model to behave in unintended ways or respond outside of ethical guidelines \cite{jiang_prompt_2023, ghosh_aegis_2024}. This is important because prompt injections can compromise the ethical integrity of AI systems, potentially leading to vulnerabilities in AI-driven applications.

Our contributions to content safety and prompt injection detection are:

\begin{itemize}
    \item We prove the intermediate hidden state between transformer layers are robust feature extractors. A penalized logistic regression classifier with the same number of trainable parameters as the size of the hidden state, with as few as 769 parameters, achieves state-of-the-art performance surpassing GPT-4o -- the presumed leader.
    
    \item We show that for both content safety and prompt injection classification tasks there exists an optimal intermediate transformer layer that produces the necessary features. With only a minimal set of training examples, the classifier generalizes extremely well to unseen examples. This is particularly valuable in cases where few high-quality examples are available, making our approach adaptable for a wide variety of custom classification scenarios.

    \item We demonstrate that the hidden states of general-purpose LLMs (Qwen 2.5 0.5B Instruct, 1.5B, and 3B) and special-purpose models fine-tuned for content safety (Llama Guard 3 1B and 8B) or prompt injection detection (DeBERTa v3 Base Prompt Injection v2), produce features achieving similar classification results. This indicates LEC generalizes across model architectures and domains. Special-purpose models require even fewer examples to surpass GPT-4o level performance on both tasks.

    \item Special-purpose content safety and prompt injection models when pruned and used as feature extractors outperform their non-pruned versions on their respective tasks.
\end{itemize}

\begin{table}[h]
\resizebox{\textwidth}{!}{%
\begin{tabular}{cccccc}
\hline
\multicolumn{6}{|c|}{Content Safety}                                                                                                                              \\
\multicolumn{1}{|c}{Source Model} &
  \begin{tabular}[c]{@{}c@{}}Trainable\\ Parameter Count\end{tabular} &
  Max F1-Score &
  \begin{tabular}[c]{@{}c@{}}F1 at \# Examples to Beat \\ Llama Guard 3 1B\end{tabular} &
  \begin{tabular}[c]{@{}c@{}}F1 at \# Examples to Beat \\ Llama Guard 3 8B\end{tabular} &
  \multicolumn{1}{c|}{\begin{tabular}[c]{@{}c@{}}F1 at \# Examples to Beat \\ GPT-4o\end{tabular}} \\ \hline
\multicolumn{1}{|c}{Qwen 2.5 0.5B Instruct} & 897                  & 0.95                 & 0.82 (5)                    & 0.82 (5)                    & \multicolumn{1}{c|}{0.87 (15)} \\
\multicolumn{1}{|c}{Llama Guard 3 1B}       & 2049                 & 0.96                 & 0.77 (15)                   & 0.77 (15)                   & \multicolumn{1}{c|}{0.83 (55)} \\
\multicolumn{1}{|c}{Llama Guard 3 8B}       & 4097                 & 0.96                 & 0.82 (15)                   & 0.82 (15)                   & \multicolumn{1}{c|}{0.82 (55)} \\ \hline
\multicolumn{1}{l}{}                        & \multicolumn{1}{l}{} & \multicolumn{1}{l}{} & \multicolumn{1}{l}{} & \multicolumn{1}{l}{} & \multicolumn{1}{l}{}    \\ \hline
\multicolumn{6}{|c|}{Prompt Injection Detection}                                                                                                                  \\
\multicolumn{1}{|c}{Source Model} &
  \begin{tabular}[c]{@{}c@{}}Trainable\\ Parameter Count\end{tabular} &
  Max F1-Score &
  \multicolumn{2}{c}{\begin{tabular}[c]{@{}c@{}}F1 at \# Examples to Beat \\ ProtectAI DeBERTa v3\end{tabular}} &
  \multicolumn{1}{c|}{\begin{tabular}[c]{@{}c@{}}F1 at \# Examples to Beat \\ GPT-4o\end{tabular}} \\ \hline
\multicolumn{1}{|c}{Qwen 2.5 0.5B Instruct} & 897                  & 0.98                 & \multicolumn{2}{c}{0.77 (5)}                      & \multicolumn{1}{c|}{0.92 (55)}  \\
\multicolumn{1}{|c}{ProtectAI DeBERTa v3}   & 769                  & 0.98                 & \multicolumn{2}{c}{0.81 (15)}                      & \multicolumn{1}{c|}{0.93 (75)}  \\ \hline
\end{tabular}%
}
\captionsetup{justification=centering}
\caption{Summary of results from LEC. Each model tested was able to surpass both their base performance as well as GPT-4o for most layers in fewer than 100 examples. The intermediate model layers attained the highest F1-scores with the fewest number of training examples across classification tasks. The above results are for the binary content safety classification task and the prompt injection classification task.}
\label{tab:summary}
\end{table}
We believe that these results offer promising indications that an LLM's inference code can be adapted to produce the necessary embeddings during the course of generating output tokens. Such an approach has the potential to reduce computational complexity to trivial levels while achieving excellent classification results on content safety and prompt injection. We discuss this more in the future work section ~\ref{sec:limitations}. 

We concur with other researchers that the various transformer layers focus on different characteristics of the prompt input \cite{valeriani_geometry_2023}. Generally, it appears that early transformer layers focus on local relationships between input tokens and later layers focus more heavily on global relationships and influencing the next token prediction. Since no part of our content safety and prompt injection classification tasks involves token generation, it seems natural to prune those layers from the model to reduce computational complexity.

Although we focused our classification exploration on content safety and prompt injection detection, these same techniques apply to any text classification task and especially those with limited training examples \cite{buckmann_logistic_2024}.

\section{Related Work}
\subsection{Language Models as Classifiers}

Previous work has demonstrated that combining transformer-based language models with penalized logistic regression on model embeddings can create effective classifiers using a small number of examples (e.g., 10 to 1,000 examples)\cite{buckmann_logistic_2024}. By using penalized logistic regression on the model embeddings, Buckmann et al. created classifiers that are robust to model size as well as quantization. This technique, called linear probing, has traditionally been used to investigate the hidden states of language models in order to understand what they represent\cite{buckmann_logistic_2024, alain_understanding_2018, cho_prompt-augmented_2023}. These works have also shown that prompting has a significant impact on the performance of these classifiers\cite{buckmann_logistic_2024}.

Our approach builds on this foundation by introducing layer-specific insights instead of only using the embeddings from the final layer of the model before the prediction head. We demonstrate that different layers are better suited for different classification tasks. This layer-focused analysis combined with model pruning enables highly accurate and efficient classification models for responsible AI focused classification tasks. 

\subsection{Language Model Pruning}

Model pruning techniques aim to reduce the computational requirements of language models by removing non-critical components of the model to reduce its overall size while maintaining performance\cite{ma_llm-pruner_2023, cheng_mini-llm_2024}. Researchers have found that in many cases up to half the model layers can be removed and that earlier layers tend to play a more important role in knowledge understanding compared to later layers\cite{gromov_unreasonable_2024}. These pruned models are typically fine-tuned to recover lost performance on tasks like zero-shot classification, generation, and question-answering\cite{ma_llm-pruner_2023, gromov_unreasonable_2024}. Research by D K, Fischer et al. shows that layer-based pruning can be used to reduce large language models by 30-50\% while maintaining nearly all of their performance as text encoders\cite{k_large_2024}. They found that using the model as a text-encoder does not require fine-tuning to recover performance, and that in some cases model performance actually increases when pruned. 

Our approach expands on this by using the hidden state from intermediate layers as input to train a classification model. Unlike other approaches, we do not need to fine-tune the model to recover performance since we find that the intermediate layers provide better performance than the original model. Our focus is on identifying which of the pruned layers create the best inputs to use for downstream classification tasks.

\subsection{Intermediate Layers}

Recent research analyzes the effectiveness of language models' intermediate layers. Each intermediate layer block encompasses 2 stages: Multi-Head Self-Attention and a Feedforward Network (MLP) with normalization occurring before and after the MLP step. After each hidden layer there is an updated contextually aware hidden state that is produced.

Valeriani et al.'s work demonstrates that in the early layers of the transformer model, the data manifold expands becoming highly dimensional, then contracts significantly in the intermediate layers, and continues to remain constant until the later layers of the model where it has a shallow second peak. Their experimentation suggests that the semantic information of the dataset is better expressed at the end of the first peak -- and therefore in the intermediate layers \cite{valeriani_geometry_2023}. 

Skean et al. found that the intermediate layers of SSMs and transformer-based models can yield better performance on a range of downstream tasks, including classification of embeddings \cite{skean_does_2024}. They show that intermediate model layers have lower levels of prompt entropy, suggesting that these layers more efficiently eliminate redundancy during training \cite{wei_diff-erank_2024}.  

Our approach further supports the claims made in these papers and goes beyond to practically demonstrate how utilizing the hidden state of the intermediate layers is beneficial for training a highly effective and computationally efficient classification model across various tasks.

\subsection{Model Explainability}

Explainability for deep learning is a critical yet challenging field of research. Due to their complexity and scale, language models are particularly opaque. Model hallucinations and the generation of harmful content are just some of the many examples that result from the lack of interpretability within a transformer model. Increased explainability offers more than the ability for researchers to improve downstream tasks, it also offers the end user clarity and confidence in the model's response.

There are several techniques to improve transformer interpretability which Luo et al. categorizes into the two broad topics of "local analysis" and "global analysis". In local analysis, researchers aim to understand how models generate specific predictions. In global analysis, researchers aim to explain the knowledge or linguistic properties encoded in the hidden state activations of a model \cite{luo_understanding_2024}. These local analysis techniques provide human-interpretable explanations of model outputs by pairing a "black box" model with a more interpretable model \cite{arrieta_explainable_2019, bollacker_extending_2019,fernandez_evolutionary_2019, papernot_deep_2018}. This allows the researcher to take advantage of the complexity of deep neural networks while maintaining some level of explainability in the output.

\begin{figure}[h]
    \centering
    \includegraphics[scale=0.7]{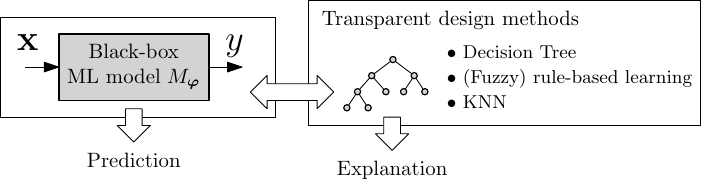}
    \caption{Visualization of a hybrid black-box model.}
    \label{fig:hybrid-blackbox}
\end{figure}

Mechanistic interpretability is a global analysis technique that seeks to explain behaviors of machine learning models' internal components. In their work, Wang et al. mechanistically interpret how GPT-2 small implements a natural language task by iteratively tracing important components back from logits, projecting the embedding space, performing attention pattern analysis, and using activation patching as part of a circuit analysis. They identify an induced subgraph of the model’s computational graph that is human-understandable and responsible for generating an output \cite{wang_interpretability_2022}. 

Although our research does not focus specifically on interpretability, it provides additional insight into the role of contextual embeddings and the hidden state between transformer layers. Our findings allow us to formulate supported hypotheses regarding why layers in some parts of the network tend to produce more effective representations for classification tasks than others. This understanding sheds light on the distinct properties of hidden layer states and their potential impact on downstream performance.

\subsection{Responsible AI Classification Tasks}

Content safety and prompt injections are two of the most well-researched and high-priority use cases related to the responsible use of Generative AI. Without effective mitigation strategies, these issues can compromise model integrity, user trust, and overall system security. Numerous methods have been developed to address these classification tasks, with public leaderboards available to benchmark their performance. Notable public leaderboards include the AI Secure LLM Safety Leaderboard on HuggingFace{\cite{noauthor_introduction_nodate}}
and the Lakera PINT Benchmark for Prompt Injection {\cite{noauthor_lakeraaipint-benchmark_2024}}. 

One detection method for prompt injections is proposed by Hung et al. where they analyze patterns in the attention heads and introduce a concept called the "distraction effect". The "distraction effect" is where select attention heads shift focus from the original instruction to the newly introduced instruction. They propose Attention Tracker, a training-free detection method that monitors attention patterns on instruction to detect prompt injection attacks{\cite{hung_attention_2024}}.

Another content safety classification approach is presented in the work of Mozes et al., where they show that LLM-based parameter-efficient fine-tuning (PEFT) can produce high-performance classifiers using small datasets (as few as 80 examples) across three domains: offensive dialogue, toxicity in online comments, and neutral responses to sensitive topics. This method offers a cost-effective alternative to large-scale fine-tuning{\cite{mozes_towards_2023}}.

Our approach distinguishes itself from these existing approaches as it prunes the LLM's hidden layers and uses just the optimal number of parameters to be the most efficient yet performant classifier for the task. Additionally, it can be implemented in two distinct ways: (1) integrated directly into the LLM's forward pass, similar to the work of Hung et al. {\cite{hung_attention_2024}},  or (2) as a separate component in the model pipeline, akin to the approach used by Mozes et al. {\cite{mozes_towards_2023}}. By offering flexibility in deployment, our method provides a versatile and scalable solution for content safety and prompt injection detection. 

\section{Experiments}

\subsection{Overview}

Our experiments explore the effectiveness of training a classifier on the hidden states of intermediate transformer layers and identify which intermediate layer(s) provide the best performance for both content safety and prompt injection classification. We compare our approach to baseline models using task-specific datasets. 

For each task, we evaluate performance against two types of baseline models, GPT-4o and a task-specific, special-purpose model. We use GPT-4o as a baseline for both classification tasks since it is widely considered one of the most capable general-purpose language models and in some cases outperforms the special-purpose models. We apply LEC to a general-purpose model and the same special-purpose model selected for the baseline. 

This setup allows us to compare three key aspects:
\begin{enumerate}
\item How well LEC performs when applied to a general-purpose model compared to both baseline models (GPT-4o and the special-purpose model). 
\item How much LEC improves the performance of the special-purpose model relative to its own baseline performance. 
\item How well LEC generalizes across model architectures and domains, by evaluating its performance on both general-purpose and special-purpose models. 
\end{enumerate}

Our experiments include models ranging from 184M to 8B parameters. We do not quantize the models, since existing research suggests that performance is largely preserved for different quantization levels\cite{buckmann_logistic_2024}. This setup provides a robust comparison of general-purpose and special-purpose models to illustrate the advantages of LEC in both responsible AI-focused classification tasks.

We select Qwen 2.5 Instruct in sizes 0.5B, 1.5B, and 3B as our general-purpose model for both content safety and prompt injection classification tasks. We select the following special-purpose models based on the classification task:

\begin{enumerate}
\item Content Safety: Meta's Llama Guard 3 1B and Llama Guard 3 8B \cite{inan_llama_2023, grattafiori_llama_2024}
\item Prompt Injection: ProtectAI's DeBERTa v3 Base Prompt Injection v2\cite{protect_ai_deberta-v3-base-prompt-injection_nodate}
\end{enumerate}

It is important to note that we do not modify the system prompts during training or evaluation to include specific instructions relevant to the classification tasks. This ensures that the LLM inference pipeline maintains its ability to be adapted to produce the necessary embeddings while generating the output tokens. We believe that this is critical for allowing classification tasks to be added to the LLM inference code without impacting computational efficiency.

We measure performance by assessing the weighted average F1-score of the baseline models and the models trained using LEC. We also examine the impact of the number of training examples on the weighted average F1-score. This method allows us to assess which layers provide the best performance for each task and how many training examples are required to achieve optimal performance.

\subsection{Experiment Implementation}

For both general-purpose Qwen 2.5 Instruct models and special-purpose models, we prune individual layers and capture the hidden state of the transformer layer to train a classification model. Our implementation uses the Python package l3prune from D K et al.\cite{k_large_2024} to load the models from their respective HuggingFace repositories and remove the LM Head. We iterate through each layer of the model, pruning a single layer each time and capturing the hidden state at the transformer layer. This allows us to understand the impact of individual layers on the task.

After pruning, we train a PLR classifier with L2 regularization on the output vector of our pruned model. Our PLR classifier uses the RidgeClassifier class from scikit-learn with $\alpha$=10. All other settings are left to their default values. Each model was run on a single A100 GPU with 220GB of memory in an Azure Databricks environment. For our GPT-4o baseline we used an Azure OpenAI deployment with version "2024-06-01". 

We used task-specific datasets, each containing 5,000 examples with 66\% allocated to training and 33\% to testing. While previous work suggests that our classifiers will only see small improvements after a few hundred examples \cite{buckmann_logistic_2024}, we randomly sampled 5,000 examples to ensure enough data diversity and minimize compute time. For each experiment, we trained multiple classifiers on a random sample of our training set from sizes 5 to 3,000. Then, we calculated the weighted F1-score on our 1,700 example test set for each unique training size, layer count, and model. To establish the baseline models' performance we calculated the weighted F1-score on the 1,700 example test set. 

\subsection{Datasets}

\textbf{Content Safety}: Our content safety dataset is designed for both binary and multi-class classification. For binary classification content is either "safe" or "unsafe". For multi-class classification, content is either categorized as "safe" or assigned to a specific fine-grained category under the broader "unsafe" category. To create a balanced dataset of safe and unsafe content, we combine two existing datasets, the "SALAD Data" dataset from OpenSafetyLab to represent unsafe content and the "LMSYS-Chat-1M" dataset from LMSYS, to represent safe content\cite{li_salad-bench_2024, zheng_lmsys-chat-1m_2024}. We randomly sample 2,500 records from each source to create a combined balanced dataset with 5,000 examples.

The full SALAD dataset contains over 21,000 unsafe messages collected from various sources and is organized into three levels of complexity. Each level contains one category for "safe" and progressively finer-grained "unsafe" categories with 6 unsafe categories at level 1, 16 at level 2, and 65 at level 3. We randomly sampled 2,500 records from the "base" set, which contains examples that have not been modified to bypass general LLM filters.

The full LMSYS dataset contains one million real-world conversations collected from 25 LLMs. We filter the dataset for English-only, first-turn messages that were not flagged as unsafe. We randomly sampled 2,500 records from the filtered dataset to use as our "safe" examples. These examples are labeled as "O0: Safe" to match the naming convention used in the SALAD dataset. 

\textbf{Prompt Injection Detection}: We use the SPML Chatbot Prompt Injection Dataset which contains 1,800 system prompts and 20,000 user prompts, from which we randomly sampled 5,000 pairs\cite{sharma_spml_2024}. In this context, a prompt injection is defined as any attempt to change the intended behavior of the AI system as defined in the system prompt. 

We chose the SPML dataset because of its diversity and complexity in representing real-world chat bot scenarios. Many other datasets for classifying prompt injection attacks are designed for general-purpose chat bots with broad instructions to not return unsafe or misleading content. Training on these simpler "no-context" prompt injection tasks often allows small models to achieve near perfect accuracy, making these datasets insufficiently challenging for evaluating our approach\cite{protect_ai_deberta-v3-base-prompt-injection_nodate}. In contrast, SPML captures nuanced and domain-specific prompt injection challenges, effectively addressing real-world challenges and providing the complexity necessary for our evaluation. Selecting real-world system and user prompts further support the possibility of using the LLM hidden state during normal LLM inference flows.

\section{Results}
\subsection{Results Summary}

Our results indicate that for both content safety and prompt injection classification tasks, using the transformer layer's hidden states with PLR classifiers consistently outperforms the baseline models, GPT-4o and the special-purpose models. Furthermore, applying LEC to the special-purpose models outperforms the models' own baseline performance by identifying the most task-relevant layer for the classification task. This ultimately results in a significant improvement in the F1-score compared to the full model's performance on the same task. Overall, we find that LEC results in improved performance across all evaluated tasks, models, and number of training examples, often achieving better performance than the baselines in fewer than 100 examples.

We find that content safety and prompt injection classification are largely driven by local features that are represented early on in the transformer network, allowing the middle layers of the model to perform better on the classification tasks than later layers. These middle layers tend to attain the highest F1-scores within the fewest training examples for content safety and prompt injection classification tasks. For both tasks, LEC enables the special-purpose models to generalize to new tasks in a similar domain using fewer than 100 training examples. We infer this capability because, to our knowledge, the base special-purpose models were not trained on the datasets used in this evaluation.

\begin{figure}[h]
    \centering
    \includegraphics[scale=0.7]{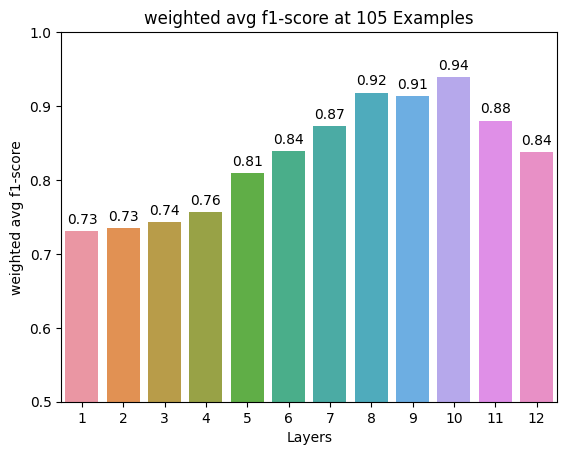}
    \caption{LEC performance at each layer of the DeBERTa-v3-Prompt-Injection-v2 model for the prompt injection task.}
    \label{fig:deberta-layers}
\end{figure}

In each model, we observe that the performance appears to mimic a continuous right-skewed function that is concave down, with one or two local maximums near 50\%-75\% of the original number of layers. This is consistent with Gromov et al.'s assertion that the layers of each model are largely dependent on the previous layer \cite{gromov_unreasonable_2024}. These findings suggest that the optimal layer for classification tasks is not the LM head or final encoding, but instead one of the intermediate layers of the model. Figure \ref{fig:deberta-layers} provides an example of the classifier performance for each layer of a model.

We also find that intermediate transformer layers tend to show the largest improvement in F1 compared to the final transformer layer when trained on fewer examples. This suggests that low-resource or low-data cases can especially benefit from LEC. In each result summary, we show the performance of LEC on 3 selected layers per model. These layers include the full model's layers, the best-performing layer, and the smallest layer that achieves similar performance to the full model.

In summary, LEC provides a more computationally efficient and better performing solution (higher F1-scores with few training examples) for these classification tasks compared to GPT-4o and the unmodified special-purpose models.

\subsection{Content Safety Results}

In this section, we present our results for both binary and multi-class classification on the content safety task. Of the baseline models used for this task -- Llama Guard 3 1B, Llama Guard 3 8B, and GPT-4o --- GPT-4o consistently achieved the highest weighted F1-scores for both binary and multi-class content safety classification. However, all the models trained using LEC consistently outperform the Llama Guard models and GPT-4o across all classification tasks, often surpassing GPT-4o's performance in as few as 20 training examples for binary classification and 50 training examples for multi-class classification.

\begin{figure}[h!]
    \centering
    \includegraphics[scale=0.35]{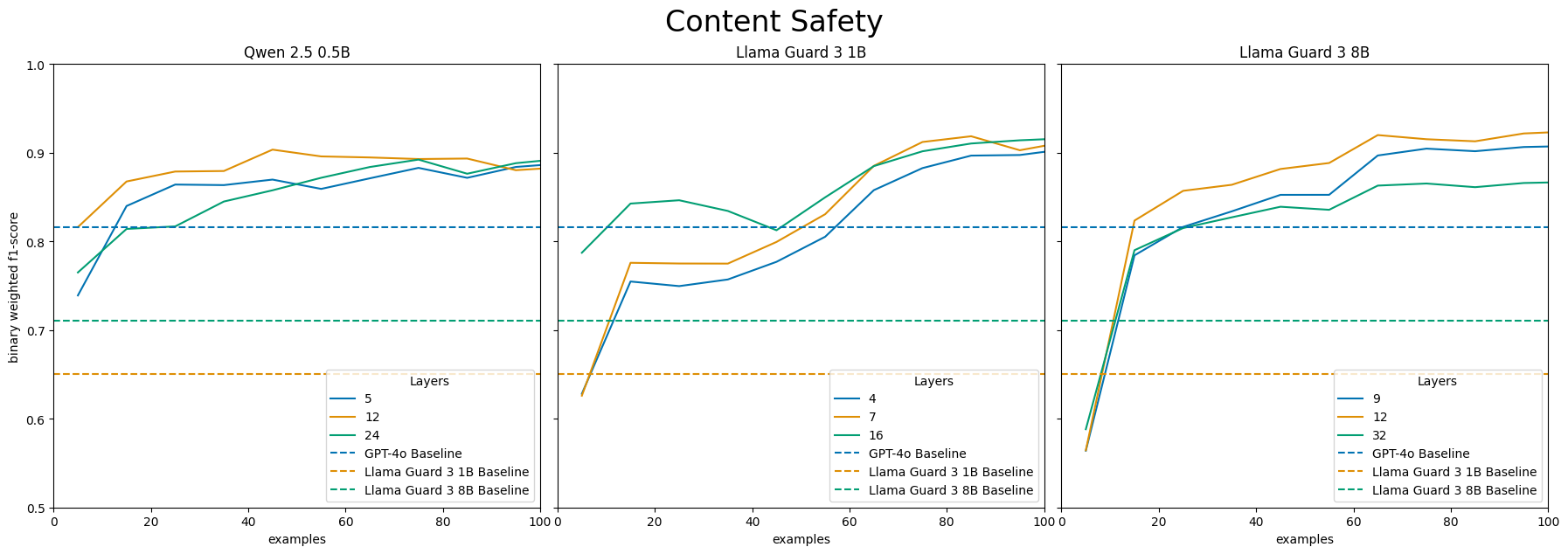}
    \caption{LEC performance of select layers on binary content safety classification for Qwen 2.5 0.5B Instruct, Llama Guard 3 1B, and Llama Guard 3 8B.}
    \label{fig:safety-results}
\end{figure}

\begin{table}[h!]
\resizebox{\textwidth}{!}{%
\begin{tabular}{|llllllll|}
\hline
Model Name &
  Layers &
  Parameter Count (B) &
  \% of Full Size &
  \begin{tabular}[c]{@{}l@{}}Max Weighted \\ F1-Score\end{tabular} &
  \begin{tabular}[c]{@{}l@{}}F1 at \# Examples to Beat \\ Llama Guard 3 1B\end{tabular} &
  \begin{tabular}[c]{@{}l@{}}F1 at \# Examples to Beat \\ Llama Guard 3 8B\end{tabular} &
  \begin{tabular}[c]{@{}l@{}}F1 at \# Examples to Beat \\ GPT-4o\end{tabular} \\ \hline
Qwen 2.5 0.5B Instruct & 5  & 0.21 & 42.65 & 0.95 & 0.74 (5)  & 0.74 (5)  & 0.84 (15) \\
Qwen 2.5 0.5B Instruct & 12 & 0.32 & 63.78 & 0.95 & 0.82 (5)  & 0.82 (5)  & 0.87 (15) \\
Qwen 2.5 0.5B Instruct & 24 & 0.49 & 100.0 & 0.95 & 0.76 (5)  & 0.76 (5)  & 0.85 (35) \\
Llama Guard 3 1B       & 4  & 0.51 & 40.94 & 0.94 & 0.75 (15) & 0.75 (15) & 0.86 (65) \\
Llama Guard 3 1B       & 7  & 0.69 & 55.71 & 0.96 & 0.77 (15) & 0.77 (15) & 0.83 (55) \\
Llama Guard 3 1B       & 16 & 1.24 & 100.0 & 0.94 & 0.79 (5)  & 0.79 (5)  & 0.83 (15) \\
Llama Guard 3 8B       & 9  & 2.49 & 33.16 & 0.95 & 0.78 (15) & 0.78 (15) & 0.83 (35) \\
Llama Guard 3 8B       & 12 & 3.14 & 41.87 & 0.96 & 0.82 (15) & 0.82 (15) & 0.82 (15) \\
Llama Guard 3 1B       & 32 & 7.5  & 100.0 & 0.94 & 0.79 (15) & 0.79 (15) & 0.83 (35) \\ \hline
\end{tabular}%
}
\caption{Content safety binary classification results for select model layers. The baseline F1-scores for Llama Guard 3 1B, Llama Guard 3 8B, and GPT-4o were 0.65, 0.71, and 0.82, respectively.}
\label{tab:safety-summary}
\end{table}

\begin{figure}[h!]
    \centering
    \includegraphics[scale=0.35]{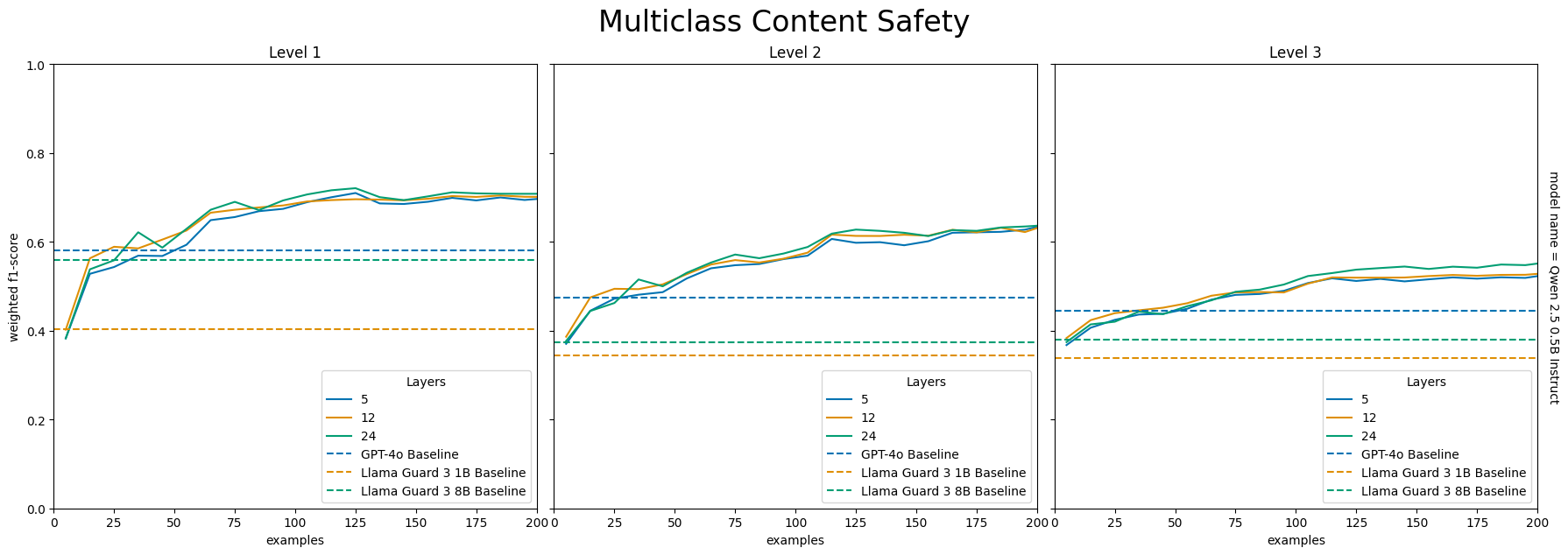}
    \caption{LEC performance of Qwen 2.5 0.5B Instruct on all three levels of the multi-class content safety dataset. }
    \label{fig:safety-results-multiclass}
\end{figure}

\begin{table}[h!]
\resizebox{\textwidth}{!}{%
\begin{tabular}{|lllllllll|}
\hline
Model Name &
  Layers &
  Parameter Count (B) &
  \% of Full Size &
  Level &
  \begin{tabular}[c]{@{}l@{}}Max Weighted \\ F1-Score\end{tabular} &
  \begin{tabular}[c]{@{}l@{}}F1 at \# Examples to Beat \\ Llama Guard 1B\end{tabular} &
  \begin{tabular}[c]{@{}l@{}}F1 at \# Examples to Beat \\ Llama Guard 8B\end{tabular} &
  \begin{tabular}[c]{@{}l@{}}F1 at \# Examples to Beat \\ GPT-4o\end{tabular} \\ \hline
Qwen 2.5 0.5B Instruct & 5  & 0.21 & 42.65 & 1 & 0.81 & 0.53 (15) & 0.57 (35) & 0.59 (55) \\
                       &    &      &       & 2 & 0.77 & 0.37 (5)  & 0.44 (15) & 0.48 (35) \\
                       &    &      &       & 3 & 0.72 & 0.37 (5)  & 0.41 (15) & 0.45 (55) \\
Qwen 2.5 0.5B Instruct & 12 & 0.32 & 63.78 & 1 & 0.82 & 0.56 (15) & 0.56 (15) & 0.59 (25) \\
                       &    &      &       & 2 & 0.78 & 0.39 (5)  & 0.39 (5)  & 0.47 (15) \\
                       &    &      &       & 3 & 0.72 & 0.38 (5)  & 0.38 (5)  & 0.45 (35) \\
Qwen 2.5 0.5B Instruct & 24 & 0.49 & 100.0 & 1 & 0.82 & 0.54 (15) & 0.59 (35) & 0.59 (35) \\
                       &    &      &       & 2 & 0.79 & 0.38 (5)  & 0.38 (5)  & 0.5 (35)  \\
                       &    &      &       & 3 & 0.73 & 0.37 (5)  & 0.41 (15) & 0.46 (55) \\
Llama Guard 3 1B       & 4  & 0.51 & 40.94 & 1 & 0.79 & 0.55 (15) & 0.57 (35) & 0.63 (55) \\
                       &    &      &       & 2 & 0.73 & 0.39 (5)  & 0.39 (5)  & 0.48 (35) \\
                       &    &      &       & 3 & 0.72 & 0.39 (5)  & 0.39 (5)  & 0.45 (55) \\
Llama Guard 3 1B       & 7  & 0.69 & 55.71 & 1 & 0.8  & 0.58 (15) & 0.58 (15) & 0.58 (15) \\
                       &    &      &       & 2 & 0.75 & 0.39 (5)  & 0.39 (5)  & 0.47 (15) \\
                       &    &      &       & 3 & 0.73 & 0.39 (5)  & 0.39 (5)  & 0.45 (45) \\
Llama Guard 3 1B       & 16 & 1.24 & 100.0 & 1 & 0.8  & 0.43 (5)  & 0.63 (15) & 0.63 (15) \\
                       &    &      &       & 2 & 0.77 & 0.43 (5)  & 0.43 (5)  & 0.5 (15)  \\
                       &    &      &       & 3 & 0.75 & 0.42 (5)  & 0.42 (5)  & 0.45 (15) \\
Llama Guard 3 8B       & 9  & 2.49 & 33.16 & 1 & 0.86 & 0.57 (15) & 0.57 (15) & 0.59 (25) \\
                       &    &      &       & 2 & 0.83 & 0.36 (5)  & 0.46 (15) & 0.48 (25) \\
                       &    &      &       & 3 & 0.77 & 0.37 (5)  & 0.41 (15) & 0.45 (45) \\
Llama Guard 3 8B       & 12 & 3.14 & 41.87 & 1 & 0.87 & 0.62 (15) & 0.62 (15) & 0.62 (15) \\
                       &    &      &       & 2 & 0.84 & 0.36 (5)  & 0.48 (15) & 0.48 (15) \\
                       &    &      &       & 3 & 0.75 & 0.37 (5)  & 0.41 (15) & 0.45 (45) \\
Llama Guard 3 8B       & 32 & 7.5  & 100.0 & 1 & 0.85 & 0.59 (15) & 0.59 (15) & 0.59 (15) \\
                       &    &      &       & 2 & 0.83 & 0.49 (15) & 0.49 (15) & 0.49 (15) \\
                       &    &      &       & 3 & 0.77 & 0.45 (5)  & 0.45 (5)  & 0.45 (5) \\ \hline
\end{tabular}%
}

\caption{Multi-class content safety classification results for select model layers across all three levels of difficulty. The baseline F1-scores for Llama Guard 3 1B across each level were 0.40, 0.34, and 0.34. The baseline F1-scores for Llama Guard 3 8B across each level were 0.56, 0.37, and 0.38. The baseline F1-scores for GPT-4o across each level were 0.58, 0.47, and 0.44.}
\label{tab:safety-multiclass}
\end{table}

For the binary classification task, all models trained using LEC outperformed the three baseline models. The hidden state of the middle layers provided the highest weighted F1-score across all general and special-purpose models. Notably, Qwen 2.5 0.5B Instruct outperformed both Llama Guard 3 baselines within 5 examples. It also surpassed GPT-4o's performance in 15 examples, attaining an F1-score of 0.87. Interestingly, using LEC, Llama Guard 3 1B surpassed its own baseline performance as well as the Llama Guard 3 8B baseline performance within 15 examples, attaining an F1-score of 0.75. Llama Guard 3 8B also surpassed all baseline models' performance with an F1-score of 0.82 in 15 examples. With additional training examples, it attained a max F1-score of 0.96. When comparing the models trained using LEC, the general-purpose and special-purpose models performed similarly, demonstrating that our approach effectively improves classification performance for both model architectures. Detailed results can be found in Figure \ref{fig:safety-results} and Table \ref{tab:safety-summary}.

For the multi-class classification task, the intermediate layers of all models using the LEC approach outperformed the GPT-4o baseline and baseline special-purpose models after being trained with fewer than 60 examples. Typical of multi-class classification problems, there is an inverse correlation between model performance and number of categories. Our findings did not suggest a clear number of layers that performed best on the multi-class content safety task, indicating that more experimentation would be needed prior to implementation based on the model in which LEC is applied to. However, by utilizing the LEC approach even very small models like Qwen 2.5 0.5B Instruct we were able to surpass the performance of baseline models such as GPT-4o for a 66 category classification problem using as few as 35 training examples. Detailed results can be found in Figure \ref{fig:safety-results-multiclass} and Table \ref{tab:safety-multiclass}.

\subsection{Prompt Injection Results}

\begin{figure}[h!]
    \centering
    \includegraphics[scale=0.4]{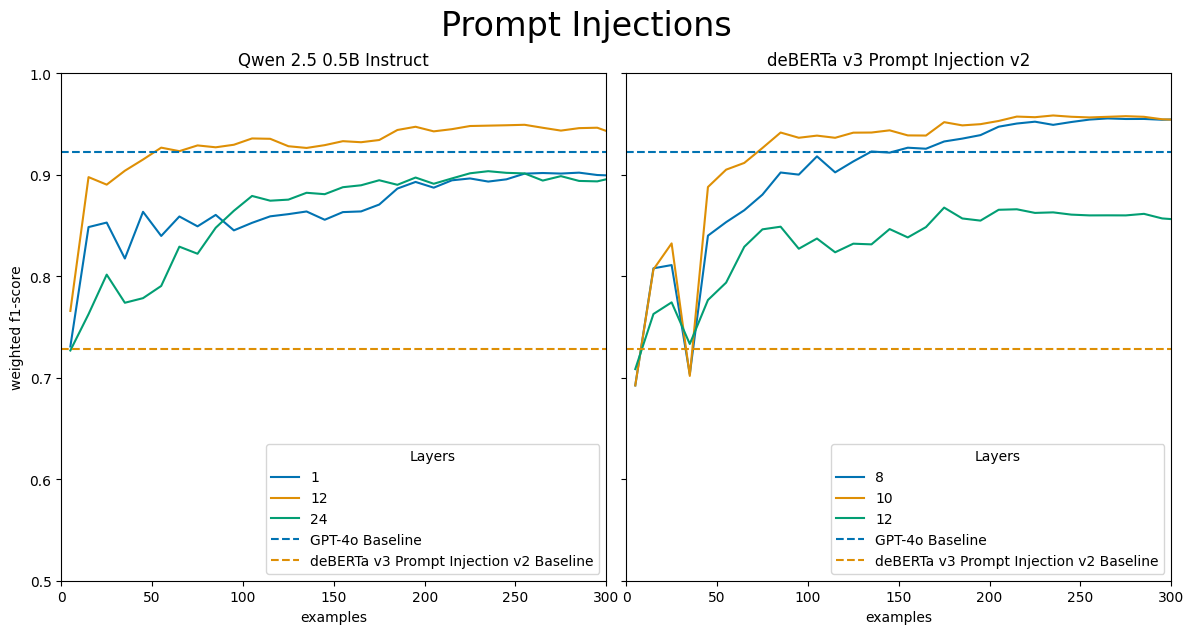}
    \caption{Performance of select layers on prompt injection classification for both general-purpose Qwen 2.5 0.5B Instruct and DeBERTa-v3-Prompt-Injection-v2.}
    \label{fig:injection-results}
\end{figure}

\begin{table}[h!]
\resizebox{\textwidth}{!}{%
\begin{tabular}{|lllllll|}
\hline
Model Name &
  Layers &
  Parameter Count (B) &
  \% of Full Size &
  \begin{tabular}[c]{@{}l@{}}Max Weighted \\ F1-Score\end{tabular} &
  \begin{tabular}[c]{@{}l@{}}F1 at \# Examples to Beat \\ ProtectAI DeBERTa v3\end{tabular} &
  \begin{tabular}[c]{@{}l@{}}F1 at \# Examples to Beat \\ GPT-4o\end{tabular} \\ \hline
Qwen 2.5 0.5B Instruct & 1  & 0.15 & 30.57 & 0.96 & 0.73 (5)  & 0.93 (900)  \\
Qwen 2.5 0.5B Instruct & 12 & 0.32 & 63.78 & 0.98 & 0.77 (5)  & 0.92 (55)   \\
Qwen 2.5 0.5B Instruct & 24 & 0.49 & 100.0 & 0.97 & 0.76 (15) & 0.94 (1000) \\
ProtectAI DeBERTa v3   & 8  & 0.16 & 84.58 & 0.98 & 0.81 (15) & 0.92 (135)  \\
ProtectAI DeBERTa v3   & 10 & 0.17 & 92.29 & 0.98 & 0.81 (15) & 0.93 (75)   \\
ProtectAI DeBERTa v3   & 12 & 0.18 & 100.0 & 0.94 & 0.73 (15) & 0.94 (2000) \\ \hline
\end{tabular}%
}
\caption{Results for select layers on the prompt injection task. The baseline F1-scores for ProtectAI DeBERTa v3 and GPT-4o were 0.73 and 0.92, respectively.}
\label{tab:injection-summary}
\end{table}

In this section, we present our results for the prompt injection classification task. We find that both general-purpose and special-purpose models trained using LEC consistently outperform all the baseline models. As can be seen in Figure \ref{fig:injection-results}, applying our approach to a general-purpose model with only 0.5B parameters (Qwen 2.5 0.5B Instruct) achieves a maximum F1-score of 0.98, surpassing DeBERTa-v3-Prompt-Injection-v2's baseline performance for all the layers within 20 examples, and outperforming GPT-4o's performance for the middle layers in 55 examples.

Qwen 2.5 Instruct surpassed DeBERTa-v3-Prompt-Injection-v2's baseline performance, F1-score of 0.73, with only 5 training examples across all model sizes (0.5B, 1.5B, and 3B) and layers. Notably, the intermediate layers achieved the highest weighted average F1-scores across model sizes. As expected, the larger Qwen 2.5 Instruct models (Qwen 2.5 1.5B Instruct and Qwen 2.5 3B Instruct) achieve similar but superior performance compared to the smallest Qwen 2.5 Instruct model, surpassing the baseline GPT-4o performance within fewer examples and attaining slightly higher F1-scores for certain intermediate layers. 

When applying LEC to DeBERTa-v3-Prompt-Injection-v2, it attained an F1-score of 0.81 for the intermediate layers in 15 examples and surpassed the baseline performance of the base  DeBERTa-v3-Prompt-Injection-v2 model. In 75 examples, layer 10 surpassed GPT-4o's performance and in 135 examples layer 12 did as well, attaining F1-scores of 0.93 and 0.94 respectively.  DeBERTa-v3-Prompt-Injection-v2's maximum F1-score for intermediate layers 8 and 10 is 0.98 which it attained with additional training examples. Detailed results can be found in Figure \ref{fig:injection-results} and Table \ref{tab:injection-summary}. 

Overall, our results show that our approach improves the performance for both general-purpose and special-purpose models on prompt injection classification. The Qwen 2.5 Instruct and DeBERTa-v3-Prompt-Injection-v2 LEC models achieve similar results in terms of F1-score and number of examples required for training. This suggests that our classification approach generalizes across both model architectures. The performance for both Qwen 2.5 Instruct and  DeBERTa-v3-Prompt-Injection-v2 LEC models approaches or exceeds GPT-4o's baseline performance at certain model sizes and layers. Due to the small size of Qwen 2.5 Instruct models tested, we believe these results demonstrate an LLM's inherent ability to extract high-quality features with significant separation so that a classifier can be trained in as few as 20 training examples. Since the DeBERTa based model also exhibits this behavior, we infer that nearly all transformer-based LLMs have this inherent ability.

\section{Conclusion}

In conclusion, our method, LEC, which uses the hidden state between intermediate transformer layers as robust feature extractors for content safety and prompt injection classification outperforms all other current methods tested including GPT-4o. The classification model is easily trained using penalized logistic regression and only a small number of training examples are needed. Most importantly, our results demonstrate that high-performing content classification is possible without modifying the LLM's weights or modifying the input prompt in any way. This classification approach has trivial computational complexity at inference time because the classifier contains the same number of parameters as the size of the LLM's hidden state. At most a few thousand new parameters are needed for the classification.

We believe that such a lightweight approach allows guardrails to be efficiently established around an LLM's input and output. This approach also unlocks the ability for other use case specific classifiers to be created. We are intrigued by the possibility that content classification can be baked into the LLM inference code providing real-time monitoring of the LLM's input and output as tokens are generated.

Our results also show that tiny LLMs can be pruned and used only as robust feature extractors for computationally efficient text classifiers. These tiny pruned models may be run virtually anywhere depending on the use case complexity.

\section{Limitations and Future Work}
\label{sec:limitations}

\textbf{Limitations}: These experiments did not fine-tune the baseline models on our datasets, we instead left them unmodified and focused on training our classifiers. We chose to use the static transformer layers to leave the possibility open that LEC could be integrated into the inference code during token generation. Fine-tuning standalone lightweight feature extractors may perform even better, but we did not explore this possibility.

Our findings are highly task dependent. More work is needed to directly compare the general ability of our method in other unexplored classification domains. Other classification domains may require more robust models, but we limited our exploration to a single linear model.

Additionally, we were unable to retrieve a small fraction of results from GPT-4o since sensitive content was blocked by built-in safety filters. Although this can affect our results since the blocked content is more likely to contain content labeled 'unsafe', this accounted for less than 1\% of our dataset in all cases. Regardless of GPT-4o's performance on these examples, our results are conclusive enough to show that our method outperforms it.

\textbf{Future Work}: LEC has numerous implications when it comes to the deployment of NLP classifiers. First, we show that models with as little as 100 million parameters and fewer than 100 training examples can make accurate predictions on a range of classification tasks. This speed and flexibility allows data scientists to test very specific use cases without significant investment in time, compute, or hardware. We also show that the hidden state of small, general-purpose models like Qwen 0.5B can be used to train a wide range of effective PLR classifiers. By using smaller models and concentrating resources into one or a handful of LM deployments, organizations can drastically reduce the amount of computational power devoted to these tasks.

Our findings also have potential implications in the field of NLP explainability. As language models grow more and more complex, there is increasing interest in interpreting and understanding their outputs \cite{zini_explainability_2023}. This is especially important for fields where automated decision-making can cause direct harm to people. Despite this, many frameworks for understanding, visualizing, or explaining transformer-based models rely on local explainability through attention layer activations on individual phrases \cite{lee_interactive_2017,vig_visualizing_2019, gholizadeh_model_2021, strobelt_seq2seq-vis_2018, hoover_exbert_2019}. In comparison, our approach uses a model that is explainable at both the local and global level. By utilizing techniques such as SHAP values \cite{lundberg_unified_2017}, we can better understand which components are most important for individual classifier predictions as well as for the model as a whole.

Our results suggest that a general-purpose model can be adapted to classify content safety violations and prompt injections while simultaneously generating output tokens. Applying LEC to a general-purpose model allows us to identify which model layers are important for each of the classification tasks. We believe it is possible to take the outputs from each of these layers, run them through their associated classification model to generate the predictions and based on the results either continue generating output tokens or stop the output from generating if a violation has occurred. Alternatively, pruning a very small language model and using its relevant layers for classification would work well and incredibly quickly, allowing for immediate identification of violations before sending the prompt inputs to a separate LLM for generation.

\clearpage
\printbibliography

\section{Appendix}
\label{sec:appendix}

\subsection{Cross-Validation Results}

\label{subsec:cross-val-results}
\begin{figure}[h]
    \centering
    \includegraphics[scale=0.7]{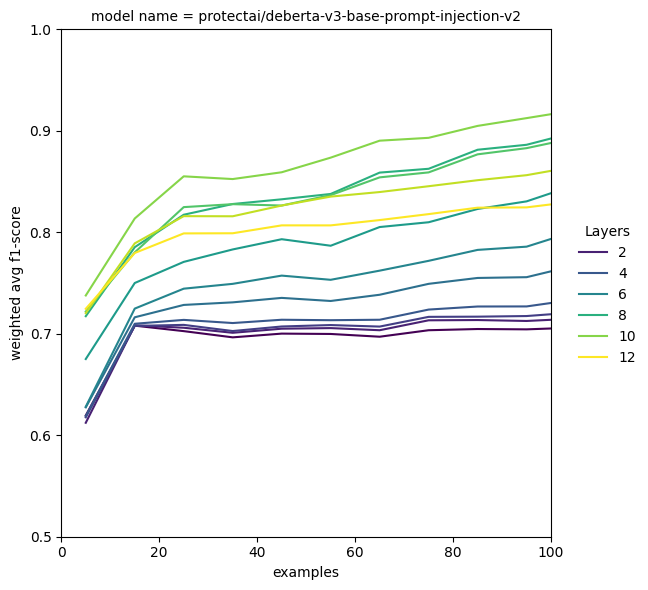}
    \caption{Performance graph of cross-validated DeBERTa LEC models on the prompt injection task.}
    \label{fig:cross-validation}
\end{figure}

\begin{figure}[h]
    \centering
    \includegraphics[scale=0.35]{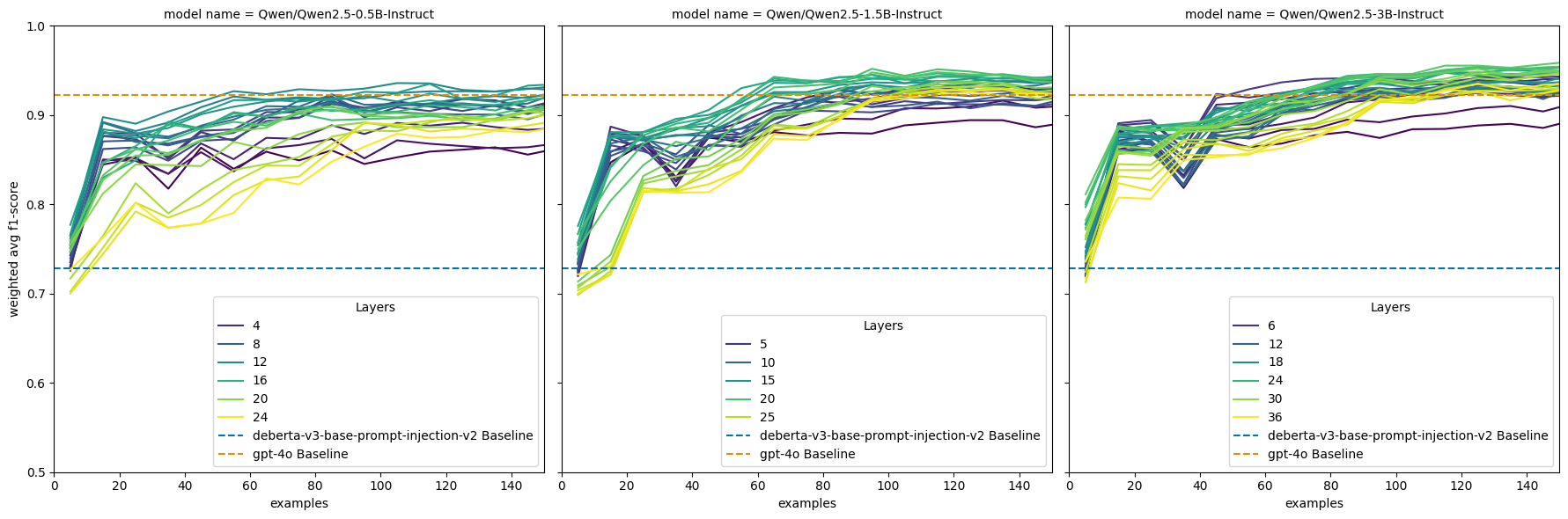}
    \caption{Full performance of each Qwen 2.5 Instruct LEC model on the prompt injection classification dataset.}
    \label{fig:pi-qwen}
\end{figure}

\begin{figure}[h]
    \centering
    \includegraphics[scale=0.35]{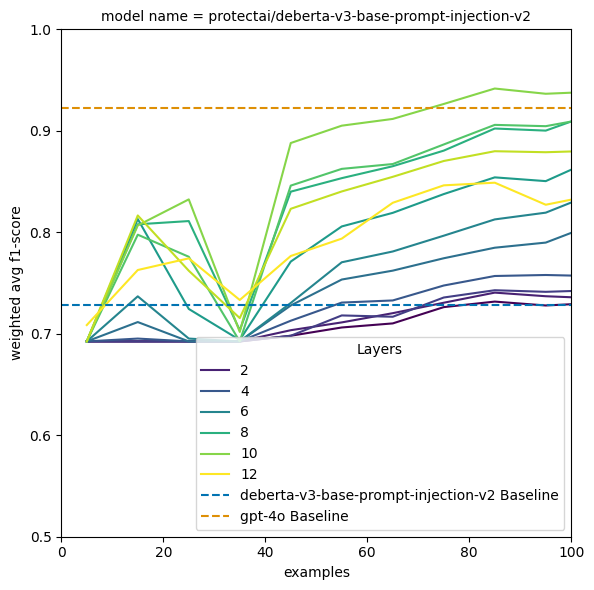}
    \caption{Full performance of the DeBERTa v3 Base Prompt Injection v2 LEC models on the prompt injection dataset.}
    \label{fig:pi-deberta}
\end{figure}

We observe that each model had large variations in performance, especially on a very small number of training examples. In the prompt injection task, there is a sharp drop in performance with a training set of around 40, which can be observed in Figures \ref{fig:pi-qwen} , \ref{fig:pi-deberta}. Interestingly, these changes were most present in the intermediate layers of each model, with the fine-tuned model having the most variability overall. Because this drop in performance is present in each model and layer trained, our assumption is that the higher variability inherent in small datasets causes sharp changes in estimated performance. We perform 10-fold cross-validation on the DeBERTa model on the prompt injection task. From Figure \ref{fig:cross-validation}, we observe that cross-validation appears to stabilize the performance across all layers of the model. We conclude that model validation metrics such as cross-validation help provide more reliable estimates of the performance of low-resource classifiers, especially when the number of examples is less than 100.

\begin{figure}[h]
    \centering
    \includegraphics[scale=0.5]{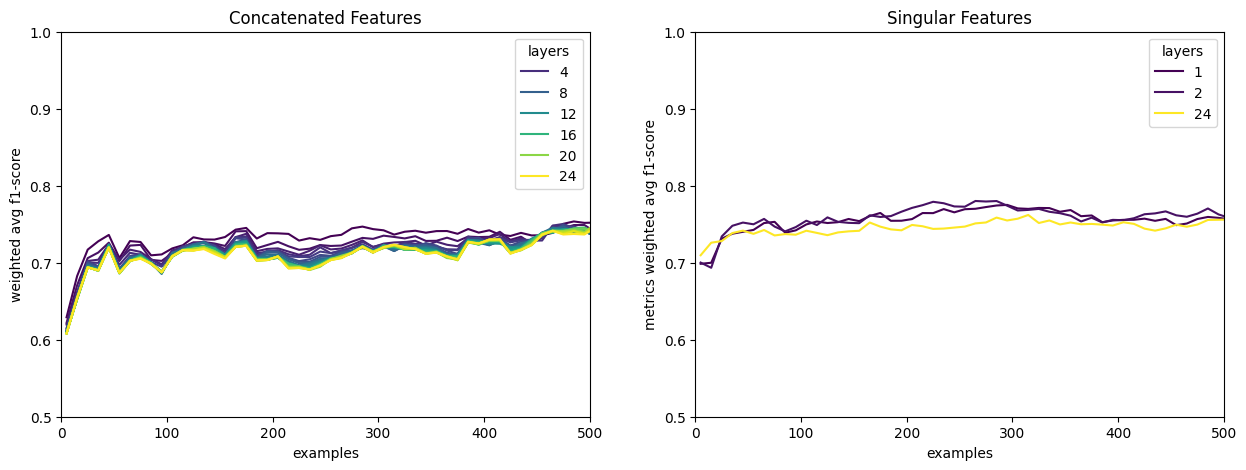}
    \caption{Performance graphs of Qwen 2.5 0.5B Instruct LEC models using concatenated layers and with a singular layer.}
    \label{fig:concatenation}
\end{figure}

\subsection{Layer Concatenation}

Since the layer representation of text can be significantly different between layers as the number of layers between them increases, we performed experiments on whether the model could benefit from using the encoding from multiple layers. In the experiment, we evaluated Qwen 2.5 0.5B using our prompt injection dataset. Rather than using a single layer's encoding, we set the features for our Ridge Classifier to be the concatenation of all previous layers. We then plotted and compared the concatenated performance to performance using single layers. From Figure \ref{fig:concatenation} and Gromov's assertion that model layers are largely dependent on the previous ones\cite{gromov_unreasonable_2024}, we conclude that layer concatenation has little to no effect on model performance.

\begin{figure}[h]
    \centering
    \includegraphics[scale=0.35]{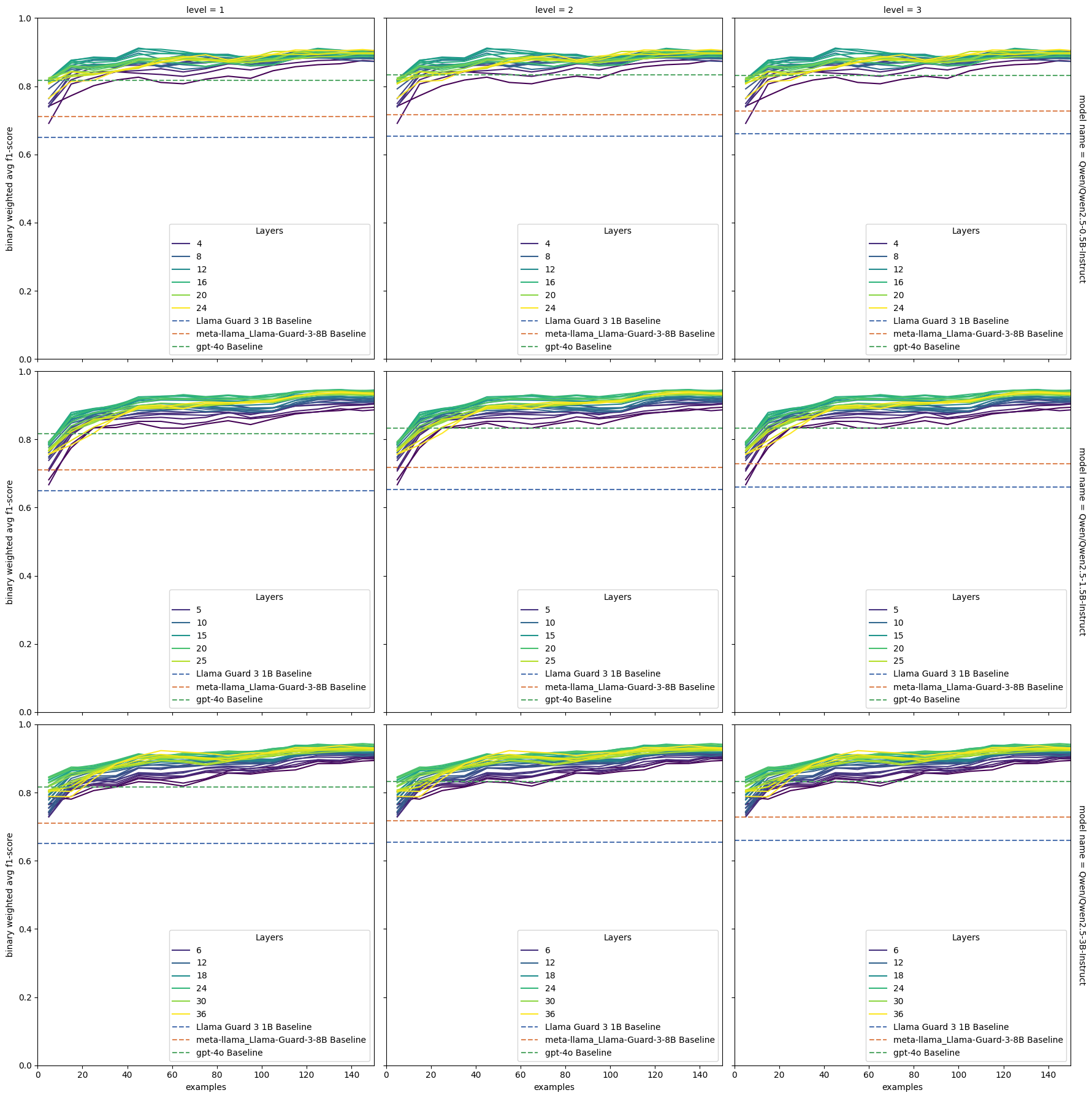}
    \caption{Full performance of each Qwen 2.5 Instruct LEC model on all 3 levels of the content safety binary classification dataset.}
    \label{fig:cs-qwen}
\end{figure}

\begin{figure}[h]
    \centering
    \includegraphics[scale=0.35]{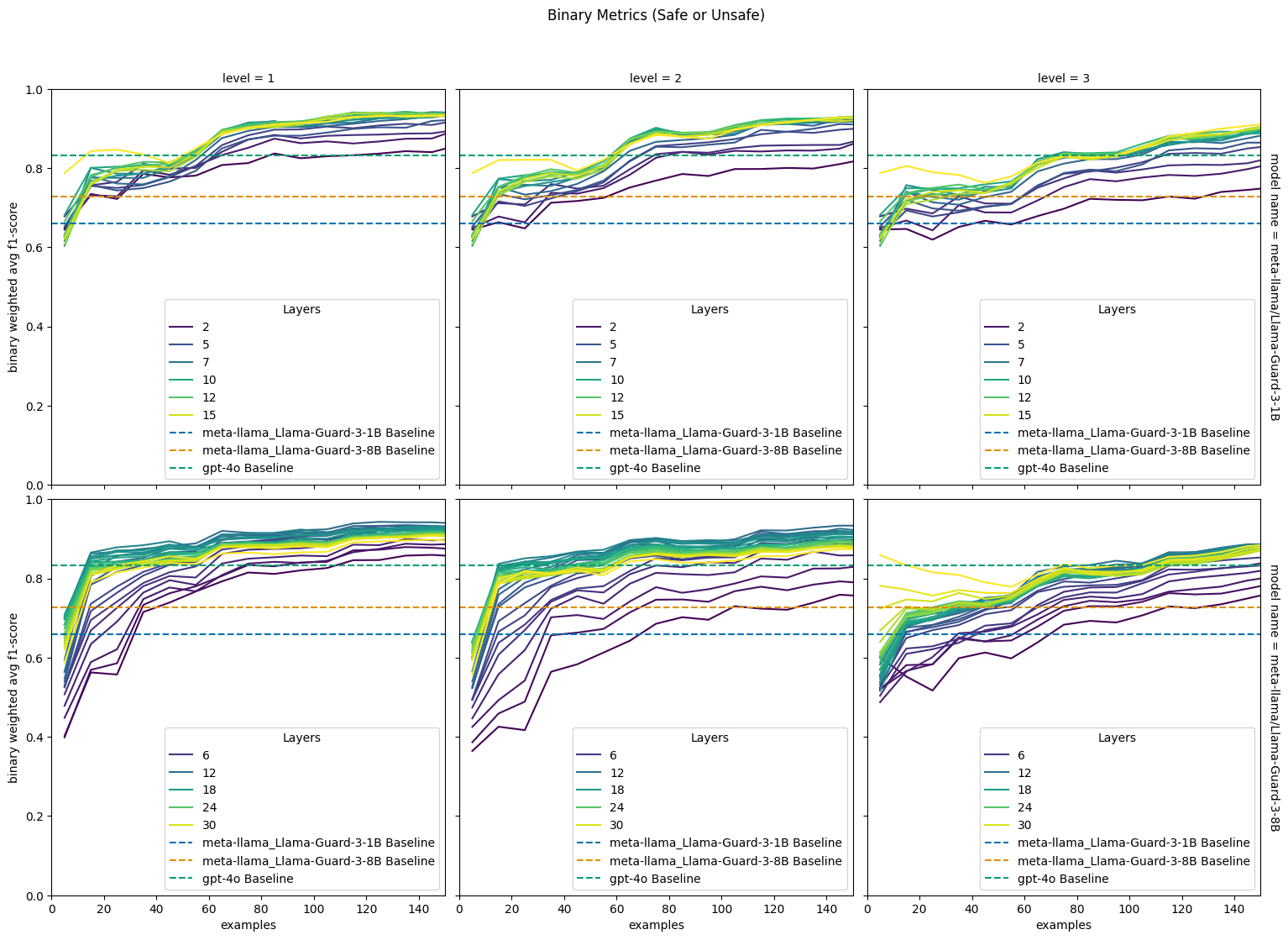}
    \caption{Full performance of each Llama Guard 3 LEC model on all 3 levels of the content safety binary classification dataset.}
    \label{fig:cs-llamaguard}
\end{figure}

\begin{figure}[h]
    \centering
    \includegraphics[scale=0.35]{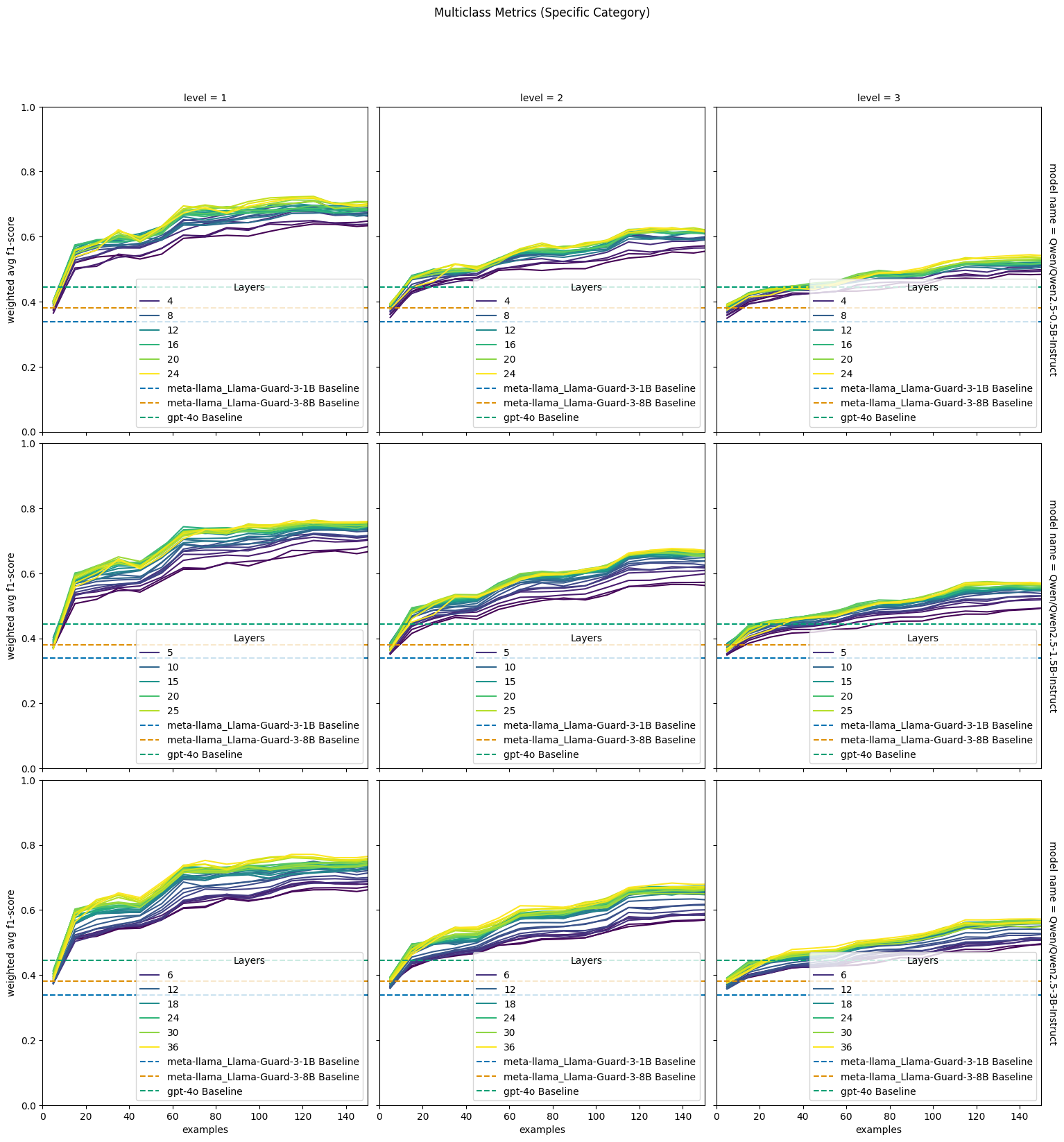}
    \caption{Full performance of each Qwen 2.5 Instruct LEC model on all 3 levels of the content safety multi-class classification dataset.}
    \label{fig:cs-qwen-multi}
\end{figure}

\begin{figure}[h]
    \centering
    \includegraphics[scale=0.35]{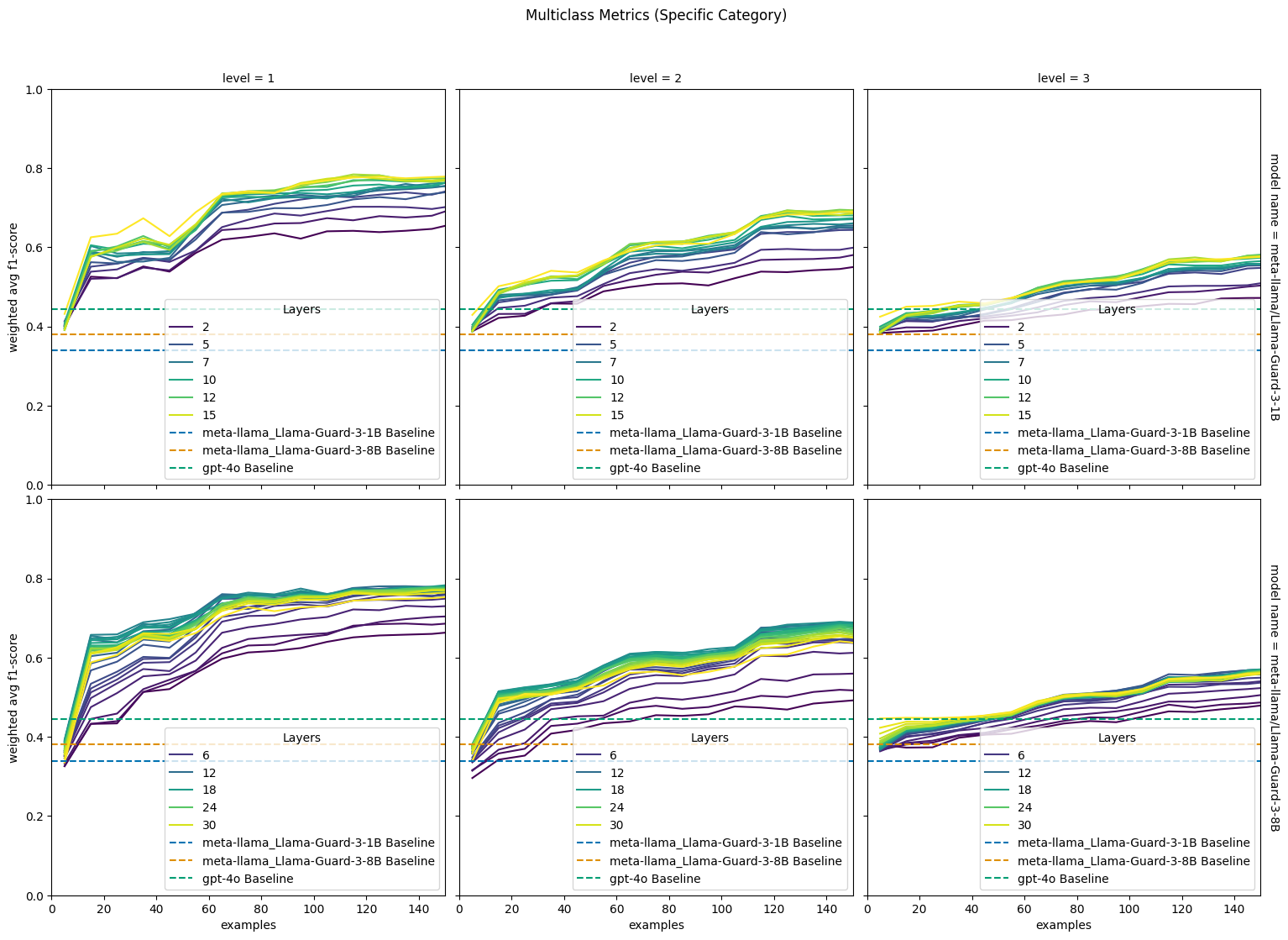}
    \caption{Full performance of each Llama Guard 3 LEC model on all 3 levels of the content safety multi-class classification dataset.}
    \label{fig:cs-llamaguard-multi}
\end{figure}

\begin{table*}[h!]
    \fontsize{9pt}{9pt}\selectfont
    \resizebox{\textwidth}{!}{%
    \renewcommand{\arraystretch}{2}
    \begin{tabular}{| p{2cm} | p{2cm} | p{6.5cm} | p{6.5cm} |}
    \hline
         \textbf{Model} & \textbf{Task} & \textbf{System Prompt} & \textbf{User Prompt}\\
    \hline
         Qwen 2.5 0.5-3B & Content Safety & You are Qwen, created by Alibaba Cloud. You are a helpful assistant. & \{User Prompt\}\\
         Qwen 2.5 0.5-3B & Prompt Injection Detection & \{System Prompt\} & \{User Prompt\}\\
    GPT-4o & Content Safety &Provide your safety assessment for ONLY THE LAST User message in the above conversation:
     - First line must read 'safe' or 'unsafe'.
     - If unsafe, a second line must include a comma-separated list of violated categories. & Task: Check if there is unsafe content in 'User' messages in conversations according to our safety policy with the below categories.
    
    <BEGIN UNSAFE CONTENT CATEGORIES>
    
    \{categories\}
    
    <END UNSAFE CONTENT CATEGORIES>
    
    <BEGIN CONVERSATION>
    
    User: \{question\}
    
    <END CONVERSATION> \\
    GPT-4o & Prompt Injection Detection & \multirow{4}{*}{\makebox[170pt][c]{\rule{50pt}{0.4pt}}} & \# Instructions
    
Classify whether the User Prompt violates the guidelines defined in the System Prompt.

\#\# System Prompt

\{System\}

\#\# User Prompt

\{User\}\\
    Llama Guard 3 1B/8B & Content Safety & \multirow{4}{*}{\makebox[170pt][c]{\rule{50pt}{0.4pt}}} & Task: Check if there is unsafe content in 'User' messages in conversations according our safety policy with the below categories.

<BEGIN UNSAFE CONTENT CATEGORIES>

\{Categories\}

<END UNSAFE CONTENT CATEGORIES>

<BEGIN CONVERSATION>

User: \{Question\}

<END CONVERSATION>

Provide your safety assessment for ONLY THE LAST User message in the above conversation:

 - First line must read 'safe' or 'unsafe'.
 
 - If unsafe, a second line must include a comma-separated list of violated categories. \\
    ProtectAI DeBERTa v3 Base Prompt Injection* & Prompt Injection Detection & \multirow{4}{*}{\makebox[170pt][c]{\rule{50pt}{0.4pt}}} &  \# Instructions
    
Classify whether the User Prompt violates the guidelines defined in the System Prompt.

\#\# System Prompt

\{System Prompt\}

\#\# User Prompt

\{User Prompt\}\\
    \hline
    \end{tabular}}
    \caption{\centering Prompts used to evaluate each model on both tasks.\newline *DeBERTa models are not chat-based, meaning the "User Prompt" is simply the text that was sent directly to the model.}
    \label{tab:promptsused}
\end{table*}

\end{document}